\documentclass[letterpaper, 10 pt, conference]{ieeeconf}  

\IEEEoverridecommandlockouts                              

\usepackage{graphics,graphicx}           
\usepackage{amsmath,amssymb}    
\usepackage{color,xspace}
\usepackage{subfigure}
\usepackage{balance}
\usepackage{cite}
\usepackage{color,soul}
\usepackage{mathtools}
\usepackage{makeidx}
\usepackage{robustindex}
\usepackage{float}
\usepackage{multirow}
\usepackage{textcomp}
\usepackage{bm}
\usepackage{trimclip}

\makeindex

\title{\LARGE \bf
Learning Extreme Hummingbird Maneuvers on Flapping Wing Robots
}

\author{Fan Fei\textsuperscript{$*$}, Zhan Tu\textsuperscript{$*$}, Jian Zhang, and Xinyan Deng
\thanks{\textsuperscript{$*$}These two authors contributed equally to this work.}
\thanks{The authors are with the School of Mechanical Engineering, Purdue University. (Email: xdeng@purdue.edu).}
}

\begin{document}

\def\AR{\text{\itshape\clipbox{0pt 0pt .32em 0pt}\AE\kern-.30emR}}

\maketitle
\thispagestyle{empty}
\pagestyle{empty}

\begin{abstract}

Biological studies show that hummingbirds can perform extreme aerobatic maneuvers during fast escape. Given a sudden looming visual stimulus at hover, a hummingbird initiates a fast backward translation coupled with a 180-degree yaw turn, which is followed by instant posture stabilization in just under 10 wingbeats. Consider the wingbeat frequency of 40Hz, this aggressive maneuver is carried out in just 0.2 seconds. Inspired by the hummingbirds' near-maximal performance during such extreme maneuvers, we developed a flight control strategy and experimentally demonstrated that such maneuverability can be achieved by an at-scale 12-gram hummingbird robot equipped with just two actuators. 
The proposed hybrid control policy combines model-based nonlinear control with model-free reinforcement learning. 
We use model-based nonlinear control for nominal flight control, as the dynamic model is relatively accurate for these conditions. However, during extreme maneuver, the modeling error becomes unmanageable. A model-free reinforcement learning policy trained in simulation was optimized to 'destabilize' the system and maximize the performance during maneuvering.
The hybrid policy manifests a maneuver that is close to that observed in hummingbirds. Direct simulation-to-real transfer is achieved, demonstrating the hummingbird-like fast evasive maneuvers on the at-scale hummingbird robot.

\end{abstract}


\section{Introduction}

Millions of years of adaptation have enabled insects and hummingbirds to possess extraordinary flight capabilities such as acrobatic aerial maneuvers including sharp turns, fast acceleration, flying backward, landing upside-down, and most impressive of all - rapid evasive maneuvers when facing threat \cite{dalton1975borne,cheng2011mechanics,cheng2016flight}. To date, much of the extraordinary abilities remain unchallenged by small scale man-made flying devices, bio-inspired flapping wing robots holds great potential to bridge this performance gap \cite{mueller2001fixed,dudley2002mechanisms}. Scientists have been decoding the basic mechanisms of flapping flight \cite{dickinson1999wing,ellington1984aerodynamics,sane2003aerodynamics}, while engineers are inspired to develop bio-inspired Flapping Wing Micro Air Vehicles (FWMAVs), of which the Nano Hummingbird\cite{keennon2012development}, the Harvard Robobee \cite{ma2013controlled}, and the Delfy \cite{karasek2018tailless} are notable examples which achieved stable hovering and maneuvering.  Beyond lift generation, flight control is a challenging task for such small scale vehicles.  Some achieve this through stroke plane and/or wing shape modulation with additional servos \cite{keennon2012development, karasek2018tailless,roshanbin2017colibri,coleman2015design}, similar to rotary wing vehicles where control is achieved through swashplate-like mechanisms, while other designs use two actuators to control the wings independently for lift generation and simultaneously achieve flight control through instantaneous wing kinematics modulation. While the former method is efficient, it lacks the ability to directly change the instantaneous wing kinematics \cite{ma2013controlled,hines2014liftoff,zhang2017design}. The latter designs with direct-driven wings hold the potential to achieve a more agile and aggressive maneuver, such as in \cite{chirarattananon2016perching}, where successful vertical perching was demonstrated.

Recent studies on hummingbirds during escape maneuver revealed how an agile flying animal performs an extremely rapid evasive maneuver \cite{cheng2016flight,cheng2016flight2}. To evoke such maneuver, researches use a visual stimulus to startle the hummingbirds. The hummingbirds responded with a stereotypical escape pattern which is a combination of pitch and yaw maneuver coupled with a linear translation. Comparing fruit fly and hummingbird maneuvers suggests that although their performance shares some similarities, the underlying dynamics are different given their differences in size, Reynolds number, and other morphological parameters. For example, Hummingbirds need significant changes in wing kinematics in order to generate the required body torque while fruit flies require only subtle wing kinematic changes due to its small size. Therefore even at its fastest maneuver, fruit fly still falls under the helicopter analogy with average wing motion control, whereas the same cannot be said for hummingbirds which require instantaneous subwingbeat wing kinematics modulation \cite{cheng2016flight}. An analysis shows that at its peak, magnificent hummingbird consumes as much as 4.5 times the power it requires for hovering. This sudden power consumption spike can only be achieved through anaerobic metabolism which insects do not possess \cite{cheng2016flight2}. This near-maximal multi-axis acrobatic maneuver epitomize the maneuverability extreme of animal fliers and makes an ideal benchmark to test the performance of man-made flapping wing robots.

\begin{figure}[!t]
\centering
\includegraphics[trim = 0mm 0mm 0mm 0mm,width=0.99\columnwidth]{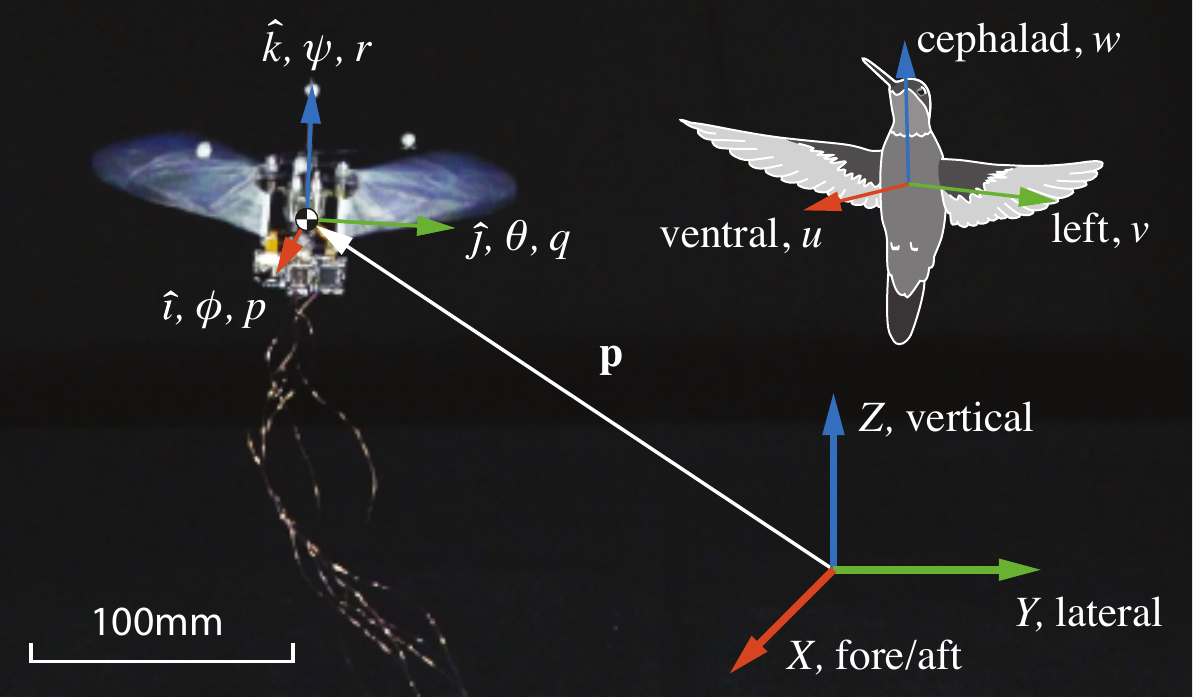}
\caption{The at-scale flapping wing robot next to an illustration of a Magnificent Hummingbird.}
\label{fig:sys}
\vspace{-0.3in}
\end{figure}

In this study, we aim to tackle this challenge by using an at-scale hummingbird robot \cite{zhang2017design} equipped with only two actuators with independently controlled wings. Specifically, we let the vehicle perform the same task as the hummingbirds escape maneuver: translating backward while completing a $180^\circ$ yaw turn as fast as it can. This is achieved through a hybrid flight control strategy which combines a model-based nonlinear controller to guarantee flight stability and a model-free maneuvering policy learns to 'destabilize' the system to maneuver. Outputs of the two control policies are added together, similar to that of the Structure Control Network\cite{pmlr-v80-srouji18a}, albeit only the maneuvering policy is learned in our case. This design choice is made because the averaged dynamics model of the system for stabilization is relatively accurate, while for extreme maneuvers the approximation error becomes unmanageable for model-based controller design \cite{deng2006flapping}. In this paper, we present the control algorithms and experimental results in hovering, figure-of-eight trajectory tracking, and most importantly - rapid evasive maneuver which is comparable to that of a real hummingbird.

\section{Model-based Control Policy}\label{sec:controller}

Modeling uncertainties in flapping flight are due to the unstable and under-actuated body dynamics, modeling and parameter uncertainty, unsteady flapping-wing aerodynamics especially during the rapid aerobatic maneuver, and damping effects induced by flapping wings, and ground effects.
To complicate the problem, parametric uncertainty coming from mechanical asymmetry, manufacture imperfections, wear and tear, operating condition change further introduce variations in vehicle dynamics. Our previous controller proposed in \cite{zhang2017geometric} has global exponential attractive property, however, it lacks the ability to deal with changing trim condition, which can slowly vary as vehicle wear and tear. While an adaptive controller can cope with system parameter changes, unmodeled dynamics can still affect the performance \cite{chirarattananon2016perching}, especially during transient. We propose a robust adaptive control law to address these challenges. Online parameter adaptation will estimate the changing system parameters, and a robust controller is used to deal with parameter estimation error and model uncertainties.

\subsection{Vehicle Dynamic Model and  Force Mapping}
The vehicle is modeled as the standard rigid body dynamics by Newton-Euler equation:
\begin{equation}
	\begin{bmatrix}
		\mathbf{F}\\
		\bm{\tau}
	\end{bmatrix} =
	\begin{bmatrix}
		m\mathbf{I}_3 & 0\\
		0 & \mathbf{J}
	\end{bmatrix}
	\begin{bmatrix}
		\ddot{\mathbf{p}}\\
		\dot{\bm{\omega}}
	\end{bmatrix}+
	\begin{bmatrix}
		0\\
		\bm{\omega}\times\mathbf{J}\bm{\omega}
	\end{bmatrix} + \bm{\Delta}
\end{equation}
where $\mathbf{F}$ and $\bm{\tau}$ are the external forces and torques applied at the center of mass of the vehicle, $\mathbf{I}_3$ is a 3$\times$3 identity, $m$ is the mass of the vehicle, $\mathbf{J}$ is the inertia matrix where the small off-diagonal terms are ignored, $\mathbf{p}=[x, y, z]^T$ is the vehicle's position in the inertial frame,  $\bm{\omega}=[p, q, r]^T$ is the body angular velocity, and $\bm{\Delta}$ is the lumped disturbance and uncertainties.

\begin{figure}[!t]
\centering
\includegraphics[trim = 0mm 0mm 0mm 0mm,width=0.99\columnwidth]{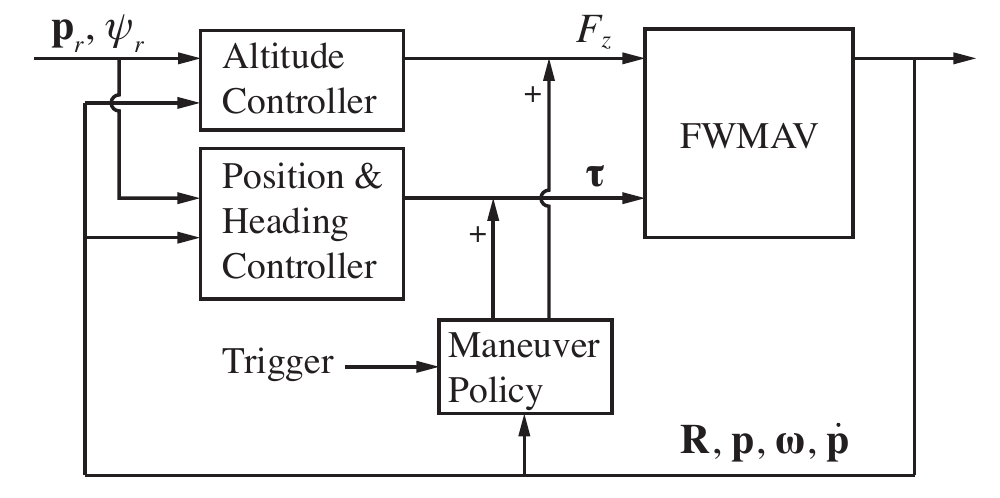}
\caption{The control structure. During normal flight, the additional maneuver policy is inactive. When given a trigger signal, the policy will generate additional torque to facilitate the evasive maneuver.}
\label{fig:control}
\vspace{-0.1in}
\end{figure}

We used the wingbeat modulation technique introduced in \cite{doman2010wingbeat} to generate the control torque and thrust, which can be approximated with linear fitting as
\begin{equation}\label{eqn:force_map}
\begin{aligned}
\tau_x &= K_\phi \delta V + \tau_{x0},& \tau_y &= K_\theta V_0 + \tau_{y0}, \\
\tau_z &= K_\psi \delta \sigma + \tau_{z0},& F_z &= K_{F_z}(u_z - V_s), \\
\end{aligned}
\end{equation}
where $\delta V$ is the amplitude difference between two motors, $V_0$ is the voltage bias, $\delta \sigma$ is the split-cycle parameter and $u_z$ is the nominal amplitude voltage of the motor drive signal. $\tau_{x0}$, $\tau_{y0}$ and $\tau_{z0}$ are the torque offsets introduced by mechanical imperfections and $V_s$ approximates the motor starting voltage.
The control structure is shown in Fig. \ref{fig:control}. The model-based control policy generates thrust and torque for stabilization, while the model-free maneuvering policy learns to 'destabilize' the system to maneuver.

\subsection{Model-based Control Design}


Here we demonstrate that, without the knowledge of the exact model and vehicle trim condition, robust performance can be achieved. The goal is to track a desired position $\mathbf{p}_r = [x_r, y_r, z_r]^T$ and heading angle $\psi_r$.
The attitude stabilization ($\phi$, $\theta$)  is merged into altitude and lateral control.

\subsubsection{Altitude Controller}
Similar to the procedure in \cite{zhang2016instantaneous}, we define a sliding surface $s_z$ for altitude control as
$s_z = \dot{e}_z + k_{z1} e_z = \dot{z} - \dot{z_{eq}}$,
where $e_z = z - z_r$ is the tracking error, $k_{z1}$ is a positive gain and $z_{eq}$ is the equivalent velocity target. The altitude error dynamics can be written as
\begin{equation}
	m\dot{s}_z = K_{u_z}u_z + \bm{\varphi}_z^T\bm{\theta}_z + \tilde{d}_z
\end{equation}
where the regressor $\bm{\varphi}_z$ and model parameters $\bm{\theta}_z$ are
\begin{align}
	\bm{\varphi}_z^T &= [-(g+\ddot{z}_{eq}), -R_{33}K_{F_z}V_s, 1]\\
	\bm{\theta}_z &= [m, 1, d_z]^T
\end{align}
$K_{u_z} = R_{33}K_{F_z}$ is the lumped input gain, $R_{33}$ is the rotation matrix element, $\tilde{d}_z = \Delta_z - d_z$ where $d_z$ is the estimated lumped nonlinear uncertainties and disturbance. We assume the unknown parameters and the uncertainty have known bounds:
\begin{equation}
\begin{aligned}
	\bm{\theta}_z \in \Omega_{\bm{\theta}_z}	 &\triangleq \{ \bm{\theta}_z : \bm{\theta}_{z_{min}} \leq \bm{\theta}_z \leq \bm{\theta}_{z_{max}} \}\\
	\tilde{d}_z \in \Omega_{\tilde{d}_z} &\triangleq \{ \tilde{d}_z : |\tilde{d}_z| \leq \delta d_z \}
\end{aligned}\label{eqn:bounds}
\end{equation}

Choose the following control law,
\begin{equation}
	K_{u_z}u_z = u_{a_z} + u_{s1_z} + u_{s2_z}\label{eqn:uz}
\end{equation}
where $u_a$ is the feedforward model compensation term, $u_{s1}$ is the stabilizing term, and $u_{s2}$ is the robust feedback term to dominate the uncertainties. They are given by
\begin{align}
	u_{a_z} &= -\bm{\varphi}_z^T\hat{\bm{\theta}}_z, & u_{s1_z} &= -k_{s1_z} s_z & u_{s2} &= -\frac{1}{4\varepsilon_z}h_z^2s_z
\end{align}
where $\hat{\bm{\theta}}_z$ is the parameter estimation, $k_{s1_z}$ is a positive gain, $\varepsilon_z$ is a sufficiently small positive number represents the approximation accuracy between the ideal sliding mode control law and the proposed method, $h_z$ is the upper bound of the modeling and parametric uncertainties. Since the error dynamics becomes
\begin{equation}
	m\dot{s}_z+k_{s1_z}s_z = u_{s2_z} - (\bm{\varphi}_z^T\tilde{\bm{\theta}}_z - \tilde{d}_z),
\end{equation}
the selected $u_{s2_z}$ satisfies $s_z(u_{s2_z} - \bm{\varphi}_z^T\tilde{\bm{\theta}}_z + \tilde{d}_z)\leq\varepsilon_z$ and $s_zu_{s2_z}\leq0$ criteria that ensures the stability of $s_z$ and avoid chattering problem in traditional sliding mode control.

With the adaptation law
\begin{equation}
	\dot{\hat{\bm{\theta}}}_z = \text{Proj}_{\hat{\bm{\theta}}_z}\left(\bm{\Gamma}_z \bm{\varphi}_z s_z\right),\label{eqn:adaptation}
\end{equation}
where $\bm{\Gamma}_z > 0$ is a diagonal adaptation rate matrix, $\text{Proj}_{\hat{\bm{\theta}}_z}(\bm{\cdot})$ is a component wise mapping as presented in \cite{zhang2016instantaneous}, $\hat{\bm{\theta}}_z$ satisfies the bounds in (\ref{eqn:bounds}) $\forall t$. Stability proof is similar to \cite{zhang2016instantaneous} and omitted for brevity.

\subsubsection{Position and Heading Controller}
Similar to the controller presented in \cite{chirarattananon2016perching}, we introduce a compound error term that includes the jerk error:
\begin{equation}
	\mathbf{e}_{xy} = \dddot{\mathbf{e}}_{\mathbf{p}} + k_{\mathbf{p}2}\ddot{\mathbf{e}}_{\mathbf{p}} + k_{\mathbf{p}1}\dot{\mathbf{e}}_{\mathbf{p}} + k_{\mathbf{p}0}\mathbf{e}_{\mathbf{p}}= \dddot{\mathbf{p}} - \dddot{\mathbf{p}}_{eq}
\end{equation}\label{eqn:exy}
where $\mathbf{e}_{\mathbf{p}} = \mathbf{p} - \mathbf{p}_r$ and $\dddot{\mathbf{p}}_{eq}$ is the equivalent third derivative target for position tracking. $k_{\mathbf{p}2}$, $k_{\mathbf{p}1}$ and $k_{\mathbf{p}0}$ are positive gains.

For heading control, the error term is defined as
$e_{r} = (r-r_r)+k_{r1}e_{\phi}^b= r - r_{eq}$,
where $r_r$ is the reference yaw velocity, $k_{r1}$ is a positive gain and $e_{\psi}^b$ is the yaw angular position error $e_{\psi} = \psi - \psi_r$ transformed into body frame.

Define a sliding surface vector, where the compound error is projected onto the body $x$ and $y$ axes.
\begin{equation}
	\mathbf{s}_{\bm{\omega}} = \begin{bmatrix}
		m(-\mathbf{e}_{xy}\cdot\hat{j})/F_z\\
		m(\mathbf{e}_{xy}\cdot\hat{i})/F_z\\
		e_{r}
	\end{bmatrix}= \begin{bmatrix}
		p + m\dddot{\mathbf{p}}_{eq} \cdot \hat{j} /F_z\\
		q - m\dddot{\mathbf{p}}_{eq} \cdot \hat{i} /F_z\\
		r - r_{eq}
	\end{bmatrix}.
\end{equation}
Thus, $\mathbf{s}_{\bm{\omega}} = \bm{\omega} - \bm{\omega}_{eq}$.

Since our current control scheme does not generate forces in the $x$ and $y$ direction, the lateral and longitudinal accelerations are related to the body rotation angles. By including jerk error, the angular velocity $\bm{\omega}$ shows up in the sliding surface variable, allowing us to directly generate control torque for position control. 

Differentiating $\mathbf{s}_{\bm{\omega}}$, the rotational error dynamics can be written as:
\begin{equation}
	\mathbf{J}\dot{\mathbf{s}}_{\bm{\omega}} = \mathbf{K}_{\mathbf{u}_{\bm{\omega}}}\mathbf{u}_{\bm{\omega}} + \bm{\varphi}_{\bm{\omega}}^T\bm{\theta}_{\bm{\omega}} + \tilde{\mathbf{d}}_{\bm{\omega}}
\end{equation}
where,
\begin{align}
	\bm{\varphi}_{\bm{\omega}}^T &= [\mathbf{I}_3, \text{diag}([-(\bm{\omega}\times\mathbf{J}\bm{\omega}) - \mathbf{J}\dot{\bm{\omega}}_{eq}]), \mathbf{I}_3]\\
	\bm{\theta}_{\bm{\omega}} &= [\tau_{x0}, \tau_{y0}, \tau_{z0},1,1,1,d_{\phi},d_{\theta},d_{\psi}]^T
\end{align}
The input gain $\mathbf{K}_{\mathbf{u}_{\bm{\omega}}} = \text{diag}([K_{\phi},K_{\theta},K_{\psi}])$, the input is $\mathbf{u}_{\bm{\omega}} = [\delta V, V_0, \delta\sigma]^T$, $\tilde{\mathbf{d}}_{\bm{\omega}} = \bm{\Delta}_{\bm{\omega}} - \mathbf{d}_{\bm{\omega}}$, where $\bm{\Delta}_{\bm{\omega}}$ is the lumped nonlinear uncertainty and disturbances, and $\mathbf{d}_{\bm{\omega}} = [d_{\phi},d_{\theta},d_{\psi}]^T$ is its estimation.

Similar to altitude control, the control law is given by
\begin{equation}
	\mathbf{K}_{\mathbf{u}_{\bm{\omega}}}\mathbf{u}_{\bm{\omega}} = \mathbf{u}_{a_{\bm{\omega}}} + \mathbf{u}_{s1_{\bm{\omega}}} + \mathbf{u}_{s2_{\bm{\omega}}}
\end{equation}
where
\begin{equation}
\begin{aligned}
	 \mathbf{u}_{s1_{\bm{\omega}}} &= -\mathbf{k}_{s1_{\bm{\omega}}} \mathbf{s}_{\bm{\omega}},
	&\mathbf{u}_{a_{\bm{\omega}}} &= -\bm{\varphi}_{\bm{\omega}}^T\hat{\bm{\theta}}_{\bm{\omega}}\label{eqn:ua_omega}\\
	\mathbf{u}_{s2_{\bm{\omega}}} &= -\frac{1}{4\varepsilon_{\bm{\omega}}} \text{diag}(\mathbf{h}_{\bm{\omega}})^2 \mathbf{s}_{\bm{\omega}}
\end{aligned}
\end{equation}
$\mathbf{k}_{s1_{\bm{\omega}}}$ is a positive diagonal gain matrix, $\mathbf{h}_{\bm{\omega}}$ is the uncertainty upper bound.

With the adaption law
$\dot{\hat{\bm{\theta}}}_{\bm{\omega}} = \text{Proj}_{\hat{\bm{\theta}}_{\bm{\omega}}}\left(\bm{\Gamma}_{\bm{\omega}} \bm{\varphi}_{\bm{\omega}} \mathbf{s}_{\bm{\omega}}\right)$,
$\mathbf{u}_{a_{\bm{\omega}}}$ can adapt to the time varying trim condition.
$\mathbf{u}_{s1_{\bm{\omega}}}$ and $\mathbf{u}_{s1_{\bm{\omega}}}$ provide the stablizing and robust feedback. The close loop stability is guaranteed.

\subsection{Discussion}
The proposed controller has three main advantages: 1) The ability to adapt to varying physical trim conditions. Mechanical imperfections of the vehicle will introduce varying torque offset \cite{chirarattananon2014adaptive, chirarattananon2016perching, zhang2017geometric}. It is impractical to repeat the trimming process before each flight test, and the controller can quickly adapt to vehicle parameter changes online. 2) Robust control can attenuate modeling uncertainties for highly complicated dynamics. 3) Unlike traditional adaptive control, by adding the robust control to deal with model uncertainties, transient performance and a large region of attraction are guaranteed.

\section{Model-free Maneuver Policy}

For the maneuvering task, we use model-free Reinforcement Learning (RL) to optimize a control policy to generate the optimal escape maneuver.
The main reason is that the cycle-averaged dynamics model of the system for stabilization is relatively accurate \cite{mellinger2011minimum,chirarattananon2016perching}, while for extreme maneuvers the approximation error becomes unmanageable for model-based controller design \cite{deng2006flapping}. During aggressive maneuvering, aerodynamic effects like FCF and FCT can affect flight dynamic drastically and could result in sub-optimal or unattainable trajectory for the vehicle.
Here, we train our RL policy in a high fidelity system simulation that includes quasi-steady flapping-wing aerodynamics, actuator dynamics, contact dynamics, joint frictions, spring nonlinearities, full aerodynamic damping, while incorporating implementation details such as sensor delay, multi-sensor fusion, and noise. The vehicle's physical parameters used in the simulation was based on the results from system identification. More details about the simulation are in \cite{fei2019flappy}.

\subsection{Problem Setup}
The maneuver policy optimization is formulated as a standard reinforcement learning setup. 
The robot's interaction with the environment is a finite-horizon Markov decision process with state space $\mathcal{S} \subseteq \mathbb{R}^{18}$, action space $\mathcal{A} \subseteq \mathbb{R}^{4}$. The simulation starts with an initial state distribution $p(s_1)$. Given a state $s_t \in \mathcal{S}$, the agent takes action $a_t \in \mathcal{A}$ according to a deterministic policy $\mathcal{\pi}_{\theta}(a_t|s_t)$. The environment generates a reward $r(s_t,a_t)$ and transition to the next state $s_{t+1}$. The state transition dynamics $p(s_{t+1}|s_t,a_t)$ is the closed loop vehicle dynamic with the controller presented in section \ref{sec:controller}. The total return is the sum of the discounted reward $\mathcal{R} = \Sigma_{i=0}^T \gamma^{i-1}r(s_i,a_i)$ at the end of the episode at time $T$. The goal is to learn a policy with parameter $\theta$ that maximizes the expected return $J = \mathbb{E}_{a_i \sim \pi_{\theta}}[\mathcal{R}]$.

To simulate the scenario presented in \cite{cheng2016flight}, the robot is first hovering at $\mathbf{p}_0=[0, 0, 0]^T$ and $\psi_0 = 0$. When given a stimulus (trigger signal), the robot will try to escape backward along the longitudinal (dorsal) direction to the new setpoint $\mathbf{p}_T=[x_T, 0, 0]^T$ and change the heading to $\psi_T = 180^\circ$ as fast as possible. Similar to \cite{cheng2016flight}, we pick $x_T = -0.21m$ which is three wing length. The magnificent hummingbird was able to complete this maneuver within 0.3 seconds. We set $T=2s$, which is long enough for the robot to finish a similar maneuver. The states and the action is defined as below
\begin{align}
	s_t &= [R_{11}, R_{12},\ ...\ , R_{33},x,y,z,p,q,r,\dot{x},\dot{y},\dot{z}]^T\\
	a_t &= [\Delta u_z, \Delta \delta V, \Delta V_0, \Delta \delta\sigma]^T	
\end{align}
The action signal is added to the output of the model-based control policy. The mathematical justification for this combination is similar to that of the SCN\cite{pmlr-v80-srouji18a}.

The reward is defined as
\begin{equation}
	r_t = (f_t+\varepsilon_f)^{-1} - \hat{i} \cdot \hat{I} + (\hat{k} \cdot \hat{K})^- + \lambda_x(x_T-x)^-
\end{equation}
where $f_t$ is the cost, $\varepsilon_f$ is a small number for numerical stability, $\hat{I}$ and $\hat{K}$ are the unit vector in the inertial frame and $\lambda_x$ is a positive scaling factor. The second term will give negative reward when the FMWAV's body $x$ axis is pointing towards the positive $X$ direction and positive reward when heading is turned towards the negative $X$ direction. The third term will generate negative reward when the body $z$ axis is pointing downwards and the last term will only generate negative reward when the agent has not reached $x_T$. These terms will encourage the agent to learn a policy that enables it to move backward away from $[0,0,0]^T$ and turn to $\psi = 180^\circ$ as fast as it can to avoid collecting negative rewards. The cost is defined as
\begin{equation}
\begin{aligned}
	f_t =& f_{s_t} + f_{c_t}\\
	=& \lambda_\mathbf{p}||\mathbf{e}_{\mathbf{p}}|| + \lambda_\mathbf{v}||\dot{\mathbf{p}}|| + \lambda_\mathbf{R}||\mathbf{e}_{\mathbf{R}}|| + \lambda_{\bm{\omega}}||\bm{\omega}||\\
	&+ \lambda_a||a_t|| + \lambda_a||\dot{a}_t||
\end{aligned}
\end{equation}
where the $\lambda$'s are positive scaling factors. The first four terms are the stability cost $f_{s_t}$, and $f_{c_t}$ is the control cost for regularization. $\mathbf{e}_{\mathbf{p}} = \mathbf{p}-\mathbf{p}_T$ is position error and $\mathbf{e}_{\mathbf{R}} = \mathbf{R} - \mathbf{R}_T$ is the attitude error wherein $\phi_T = 0$, $\theta_T = 0$ and $\psi_T = 180^\circ$.


\subsection{Training}

As a continuous control problem, we use the off-policy actor-critic algorithm Deep Deterministic Policy Gradient (DDPG) \cite{lillicrap2015continuous} as the training algorithm. Standard fully-connected MLPs (multilayer perceptrons) were used as function approximators. We use 2 hidden layers of 32 hidden units for the actor network and 2 hidden layers of 64 hidden units for the critic network. Both networks use tanh for hidden and output activation. The simulation is running at 10kHz and the control loop at 500Hz to be consistent with the physical robot. Each epoch runs for 10,000 samples which correspond to 20 seconds or at least 10 episodes. The implementation is based on \cite{duan2016benchmarking} with the same hyperparameters as \cite{lillicrap2015continuous} except the reward is scaled by 0.05 to keep the total return in a reasonable range. If the agent completes the maneuver successfully, it will collect positive reward near $\mathbf{p}_T$ until the end of the episode.


To help simulation-to-real transfer, we use the dynamics randomization technique\cite{peng2018sim}. We inject randomness into the physical parameters such as mass/inertia and motor parameters, also mechanical trims such as spring stiffness, wing midstroke position etc., as well as adding noise to sensor and actuation signal \cite{tobin2017domain}.


\section{Experimental Results}
The FWMAV experimental platform is described in details in \cite{zhang2017design}. The vehicle has a wingspan of 167.9 mm, weights 12.1 grams, and is able to generate up to 26 grams of lift, 2.71N/mm roll, 1.12N/mm pitch and 0.53N/mm yaw torques. Two motor drivers, a MARG (Magnetic, Angular Rate and Gravity) sensor and a microcontroller are onboard. For attitude feedback, sensor fusion algorithm presented in \cite{tu2018realtime} is used. The sensor fusion and control algorithms are running onboard at 500 Hz and motor commutation at 10kHz.  A VICON motion caption system with 6 cameras running at 200 Hz is used to provide position feedback. A simplified version of the algorithm presented in \cite{lynen2013robust} is used to fuse VICON and accelerometer data at 500Hz to deal with VICON data transmission delay and provide fast and robust position feedback.

\subsection{Controller Verification}
\begin{figure}[!h]
\centering
\includegraphics[trim = 0mm 0mm 0mm 0mm,width=0.9\columnwidth]{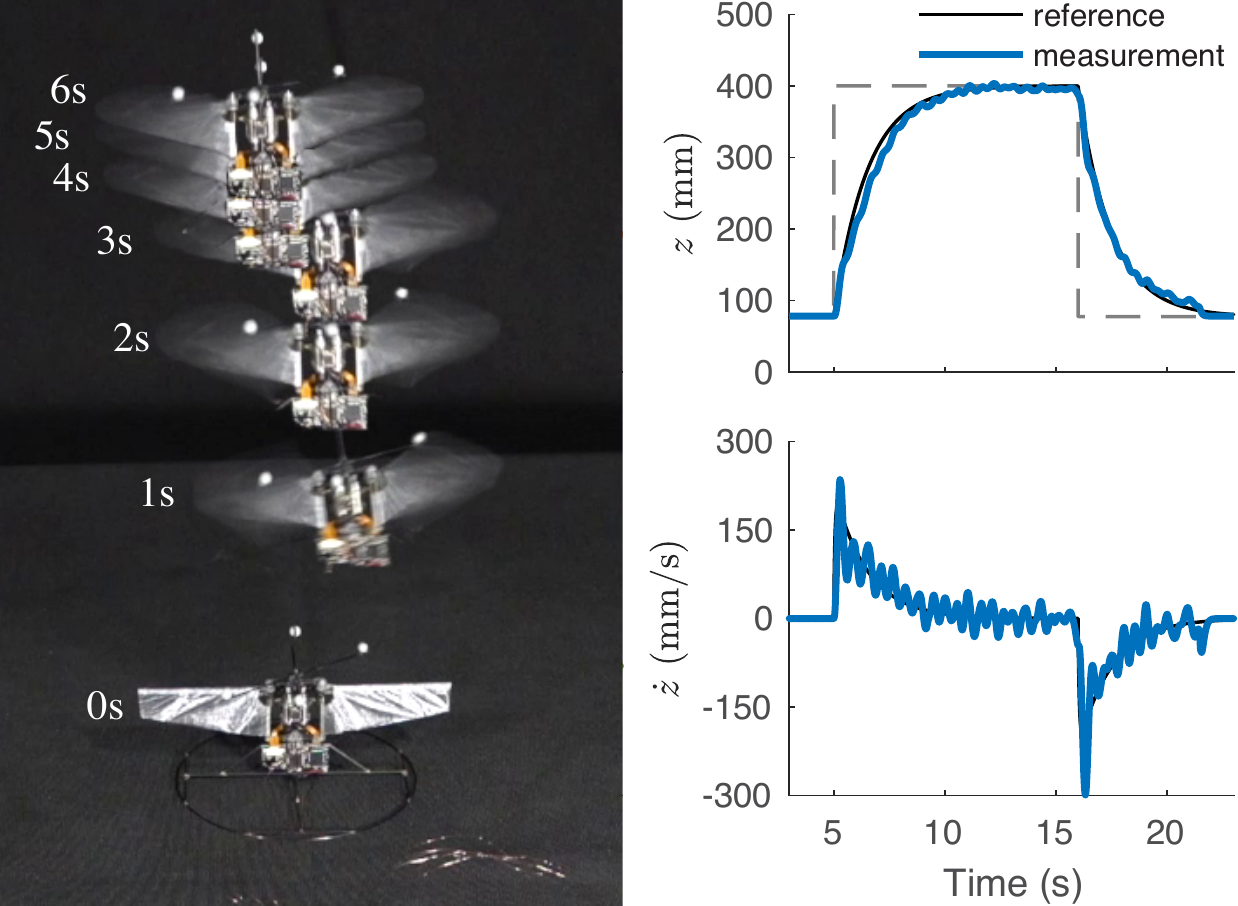}
\caption{Hovering and altitude tracking. The vehicle is commanded to takeoff and hover at a height of 40 cm during the first 11 seconds, then land afterwards. The motors shut off when position and altitude errors were smaller than 10mm.}
\label{fig:z_tracking}
\end{figure}

First, for hovering and altitude tracking, a filter is used to generate the $z$, $\dot{z}$ reference with prescribed performance requirement when giving a setpoint. The step response of altitude tracking is shown in Fig. \ref{fig:z_tracking}. The $z$ velocity shows some oscillation, but the transient and final position tracking error is very small. The vehicle performed very well during take-off, tracking, and landing.

\begin{figure}[!h]
\centering
\includegraphics[trim = 0mm 0mm 0mm 0mm,width=0.95\columnwidth]{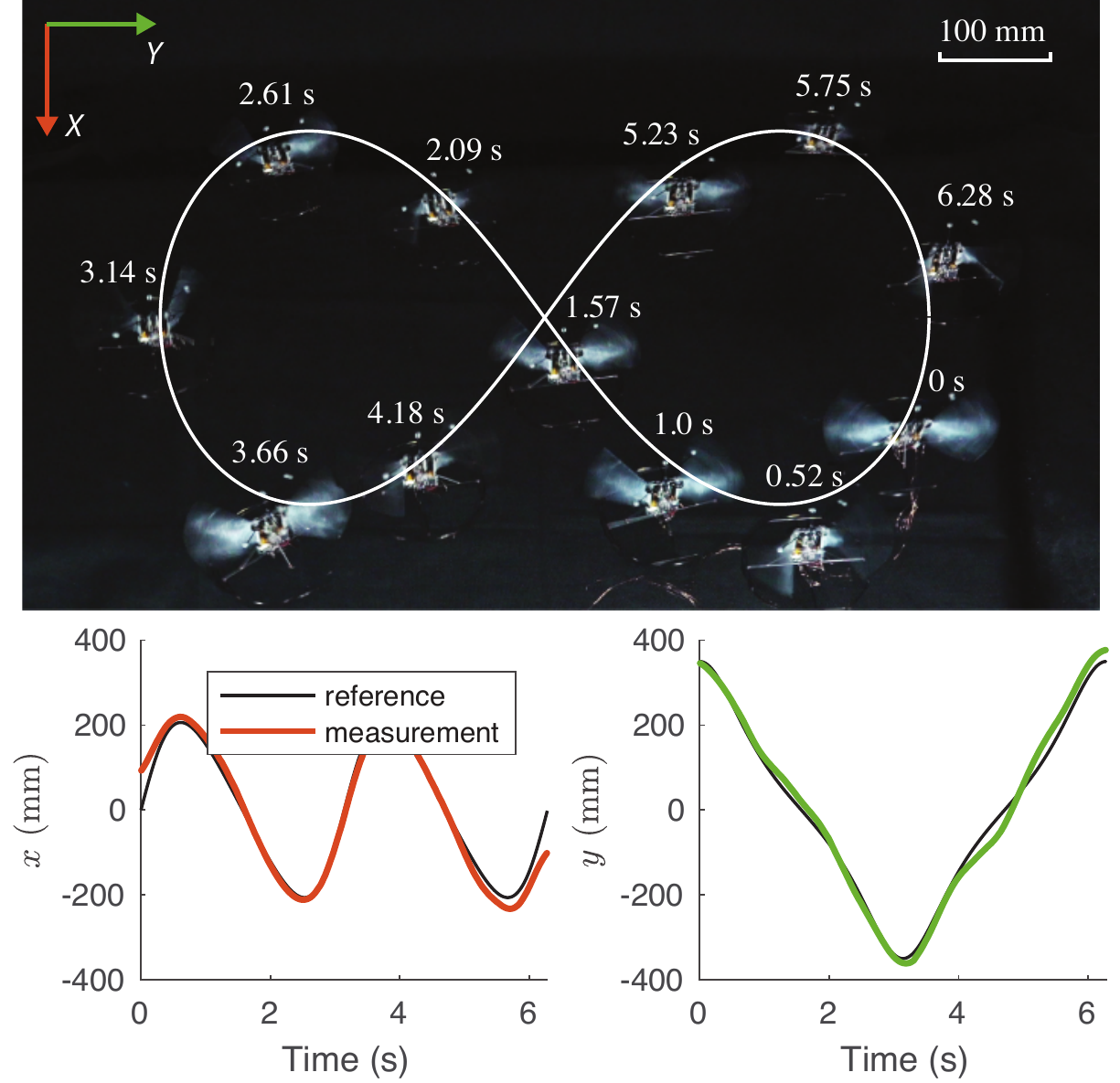}
\caption{Top: A composed image showing robot tracking a figure of 8 trajectory in the $X$-$Y$ plane. The image is transformed so the true lateral position and error relative to the reference trajectory is shown. Bottom: Lateral and longitudinal position tracking of a figure of 8 trajectory.}
\label{fig:figure_8_composed}
\vspace{-0.1in}
\end{figure}

Next, trajectory tracking using a figure-of-eight path is illustrated. A fourth order filter is used to generate $\mathbf{p}_r(t)$ when the command waypoints are given, so its third derivative is Lipschitz during implementation. Fig. \ref{fig:figure_8_composed} shows the trajectory tracking in $X$ and $Y$ direction and demonstrates the actual trajectory in the horizontal plane.

\begin{table}[!b]
\vspace{-0.05in}
\caption{Controller tracking error}
\begin{center}
\begin{tabular}{ c | c c c | c c c}
\multirow{2}{*}{$(mm)$} & \multicolumn{3}{c|}{Transient} & \multicolumn{3}{c}{Steady-state}\\
& Mean & RMS & Max & Mean & RMS & Max\\
 \hline
 $x$ & -3.81 & 21.92 & 95.20 & 0.47 &18.67 & 51.80\\
 $y$ & -0.97 & 27.20 & 62.01 & 6.12 &14.63 & 32.49\\
 $z$ & 5.22 & 11.33 & 41.52 & -0.63 & 3.95 & 13.75
\end{tabular}
\end{center}
\label{tab:error}
\vspace{-0.1in}
\end{table}

The transient and steady-state (hovering) tracking errors are evaluated over 10 different flights ($123.7s$). The mean error, error root mean square (RMS) and absolute maximum error are shown in table \ref{tab:error}. The result proves that the controller is able to maintain the position mostly within one wing length ($70mm$) during hovering and maneuvering flight, with maximum $x$ position error went slightly larger during maneuvering. The error RMS is very close between transient and steady state, which shows the controller can adapt to unknown disturbances and model uncertainty quite well. The maximum altitude control error is within one mean chord length ($21.2mm$) at steady-state and two mean chord during takeoff and landing.

\subsection{Evasive Maneuver}
\begin{figure*}[!t]
\centering
\includegraphics[trim = 0mm 0mm 0mm 0mm,width=0.95\textwidth]{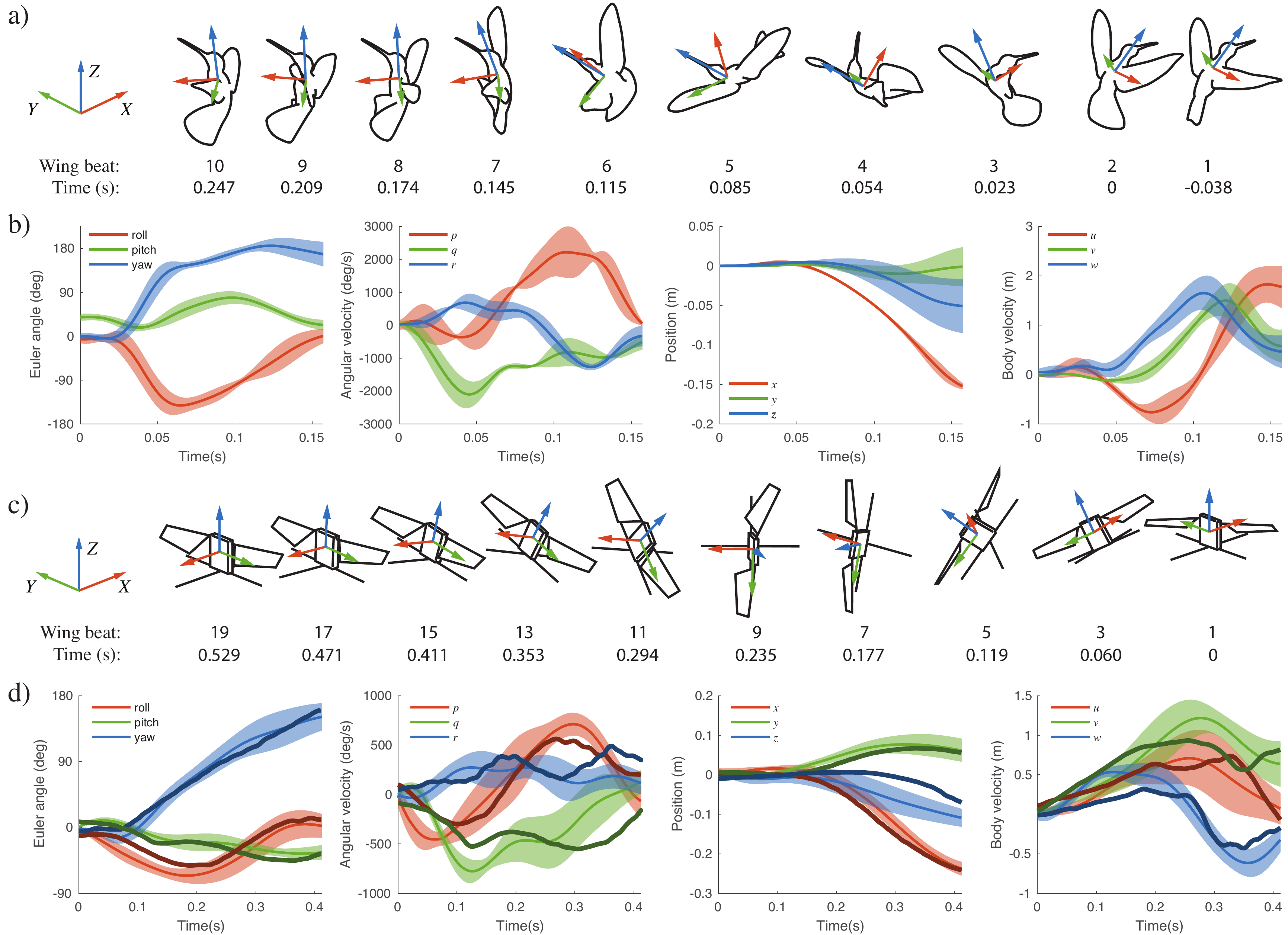}
 \caption{Comparison of the evasive maneuver of hummingbird and robot. a). Evasive maneuver sequence of a Magnificent Hummingbird[3]. b). Body kinematics of magnificent hummingbird evasive maneuver averaged over three flights. c). Evasive maneuver sequence of the FWMAV robot with optimized policy. d). Body kinematics of FWMAV robot averaged over ten flights and sample experimental data superimposed on top.}
\label{fig:mag_vs_sim}
\vspace{-0.1in}
\end{figure*}

The morphological parameters of a magnificent hummingbird studied in \cite{cheng2016flight} and our vehicle are compared in table \ref{tab:mag_vs_fwmav}. The wing morphology (wing length $R$, aspect ratio $\AR$ and second moment of wing area $\hat{r}_2$) and flapping frequency $f$ during evasive maneuver are very similar. The optimized policy after 500 episodes of training manifests a behavior which shares similarity to that observed in hummingbirds. With limited wing actuation and degrees of freedoms, and 1.5 times the body mass of the real animal, the robot completes the escape maneuver in 20 wingbeats, versus 10 wingbeats observed in the hummingbird. The comparison of body kinematics of the Magnificent Hummingbird and the hummingbird robot in simulation is shown in Fig. \ref{fig:mag_vs_sim}.

\begin{table}[!t]
\vspace{-0.05in}
\caption{Hummingbird \& FWMAV characteristics}
\begin{center}
\begin{tabular}{ c | c c c c c}
& $m(g)$ & $f(Hz)$ & $R(mm)$ & $\AR$ & $\hat{r}_2$\\
\hline
Hummingbird & 7.9 & 31.6 & 77 & 7.9 & 0.49\\
 FWMAV & 12.1 & 34 & 70 & 6.6 & 0.53
\end{tabular}
\end{center}
\label{tab:mag_vs_fwmav}
\vspace{-0.1in}
\end{table}

\begin{figure*}[!t]
\centering
\includegraphics[trim = 0mm 0mm 0mm 0mm,width=0.95\textwidth]{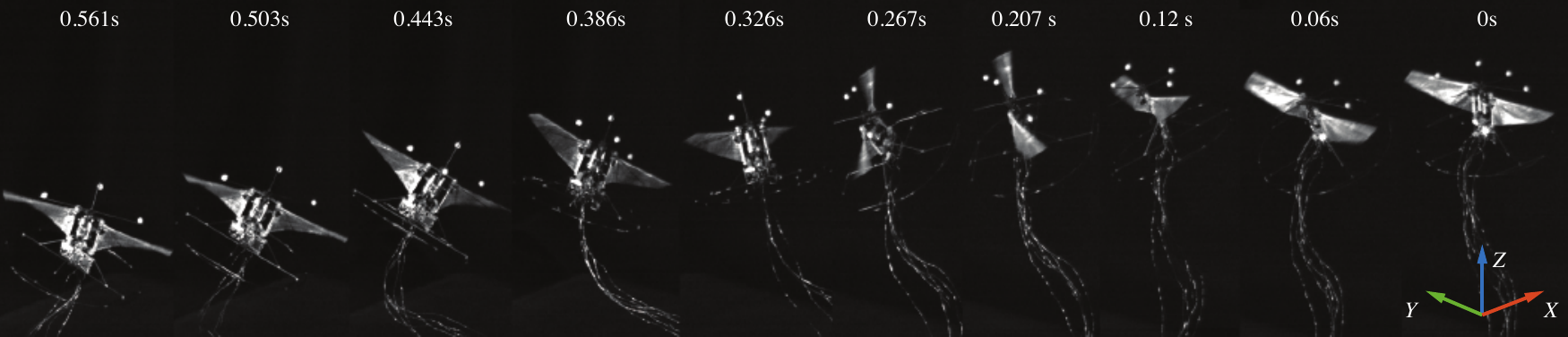}
\caption{Time sequence of the FWMAV escape maneuver showing every other wingbeat. The maneuver follows positive roll and negative yaw given the vehicle trim condition. The trajectory is plotted in Fig. \ref{fig:mag_vs_sim}d.}
\label{fig:99_seq}
\vspace{-0.2in}
\end{figure*}

The hummingbird completes the evasive maneuver in three stages as seen in Fig. \ref{fig:mag_vs_sim}. Similar movement was also observed on hawkmoth \cite{cheng2011mechanics}. During the first stage (wingbeat 1-4), it initiates a pitch-up (negative $y$) and backward (negative $X$) translation. At the end of this stage, the body $z$ axis is almost horizontal and points to the direction of escape. During the second stage (wingbeat 4-7), the hummingbird maintains its horizontal pose and starts to yaw, which in turn changes the body $x$ axis from pointing upwards to downwards. The last stage is the recovering stage, where the hummingbird returns to hover pose and decelerates.

The overall motions for both the hummingbird and the robot follows the same pattern: backward and pointing body horizontal toward the direction of escape, yaw turn, then pointing body upwards again. For comparison, we break down robot maneuver similar to that of the hummingbird. The vehicle starts its maneuver with a combined negative roll, negative pitch and positive yaw motion. At the end of this stage (wingbeat 1-5), the body $z$ axis is almost horizontal just like the hummingbird, however, the body positive $x$ axis (ventral) is already pointing up and to the side. At the end of the first stage, the vehicle gained a large negative pitch velocity. In the second stage (wingbeat 5-13), since the vehicle has relatively weak yaw torque, it cannot complete the $180^\circ$ yaw turn in time and point the $x$ axis downwards. Instead, the vehicle will leverage its ability to generate strong roll and pitch torque, initiate a combined roll and pitch motion, and brings the $z$ axis pointing upwards again. During the last stage, the vehicle stabilizes and continues to generate yaw torque until the ventral axis aligns with the negative $X$ axis. A time sequence of the experimental result is shown in Fig. \ref{fig:99_seq}. Since the real vehicle generates a net negative yaw torque given the physical trim condition, the maneuver mirrors that observed in the simulation.

The robot's ability to generate yaw torque is the weakest among the three axes due to its being severely underactuated with only two actuators. Although they share many similarities, the optimal maneuver is slightly different from that of the real hummingbird which has many groups of power and steering muscles. Given their different capabilities to generate yaw torque, the hummingbird was able to complete $180^\circ$ yaw turn in just 4 wingbeats, whereas the robot keeps on generating yaw torque throughout the entire maneuver. Furthermore, the hummingbird can change its wing kinematic to generate force in $x$ direction to decelerate, whereas the robot relies on pointing the thrust in the $X$ direction to decelerate. Nevertheless, our result is still very promising as it suggests that a flapping wing vehicle can produce animal like extreme maneuver even with severely limited wing actuation.

\section{Conclusions}
\vspace{-0.1in}
In this work, we demonstrated that a hummingbird-sized flying robot is capable of maneuvering like its natural counterpart. Despite the severely limited actuation (2 actuators only) and the limited degree of freedom in wing motion, the vehicle can still achieve agile movements. This is enabled by a hybrid flight control strategy which combines a model-based nonlinear controller to guarantee flight stability and a model-free reinforcement learning maneuvering policy learns to 'destabilize' the system to maneuver. Sim-to-real transfer results show that, even with limited wing actuation on the robot, it can still perform an aggressive movement that is similar to the Magnificent Hummingbird.



\balance

\bibliography{all}
\bibliographystyle{IEEEtran}

\end{document}